\documentstyle[alltt,11pt]{article}

\def\ar{\rightarrow}
\def\set#1{\{#1\}}
\def\armath{\(\rightarrow\)}

\def\myqquad{{\hspace*{2em}}}

\title{Charts, Interaction-Free Grammars, and the\\ Compact
  Representation of Ambiguity \thanks{To be published in IJCAI-97
    Proceedings (see {\tt http://www.ijcai.org} for further information).}}

\author{Marc Dymetman\\
Rank Xerox Research Center\\
6, chemin de Maupertuis\\
38240 Meylan, France\\[1ex]
{\tt Marc.Dymetman@grenoble.rxrc.xerox.com}}

\date{January 1997}

\begin{document}

\maketitle

\begin{abstract}

  Recently researchers working in the LFG framework have proposed
  algorithms for taking advantage of the implicit context-free
  components of a unification grammar \cite{Maxwell96}.  This paper
  clarifies the mathematical foundations of these techniques, provides
  a uniform framework in which they can be formally studied and
  eliminates the need for special purpose runtime data-structures
  recording ambiguity. The paper posits the identity: {\em Ambiguous
    Feature Structures = Grammars}, which states that (finitely)
  ambiguous representations are best seen as unification grammars of a
  certain type, here called ``interaction-free'' grammars, which
  generate in a backtrack-free way each of the feature structures
  subsumed by the ambiguous representation. This work extends a line
  of research \cite{bl89,Lang94} which stresses the connection between
  charts and grammars: a chart can be seen as a specialization of the
  reference grammar for a given input string. We show how this
  specialization grammar can be transformed into an interaction-free
  form which has the same practicality as a listing of the individual
  solutions, but is produced in less time and space.\\[5mm]
\end{abstract}

\nocite{*}

\section{Introduction}

Chart-parsing is a well-known technique for representing compactly the
multiple parses of a CF grammar. If $n$ is the input string length,
the chart can register all the parses in $O(n^3)$ space, although
there may be an exponential number of them. Each parse can then be
recovered in linear time from the chart.

Chart-parsing can be extended to CF-based unification grammar
formalisms such as LFG or DCGs. In this case, however, the valid
parses of the input string cannot be recovered so easily from the
chart. Interaction between feature structures can in theory lead to
np-complete complexity: printing the first valid parse may require
exponential time.

Such complex interactions are however rare in natural language. There
is growing agreement among researchers about the ``mild
context-sensitiveness'' of natural language
\cite{josh85,MST::Vijay-ShankerW1994}. This means that NL grammars
deviate only minimally in complexity from the context-free class.
Thus, although NL grammars written in a unification formalism may
appear superficially to be highly context-sensitive, most of the
interactions between features tend to be local, so that many of the
unification constraints could in principle be replaced by fine-grain
context-free rules.

Recently researchers working in the LFG framework have proposed
algorithms for taking advantage of the implicit context-free
components of a unification grammar. Several related algorithms have
been proposed, which rely on a notion of ``disjunctive lazy copy
links'', a form of information propagation in feature structures which
is only triggered in case of possible interactions between features
\cite{Maxwell96}.

%
%
%

This paper clarifies the mathematical foundations of these techniques,
provides a uniform framework in which they can be formally studied and
eliminates the need for special purpose runtime data-structures
recording ambiguity.  The paper posits the identity:
\begin{quote}{\bf
  Ambiguous Feature Structures = Grammars},
\end{quote}
which states that (finitely) ambiguous representations are best seen as
unification grammars of a certain type, here called
``interaction-free'' grammars, which generate in a backtrack-free way
each of the feature structures subsumed by the ambiguous
representation. This work extends a line of research
\cite{bl89,Lang94} which stresses the connection between charts and
grammars: a chart can be seen as a specialization of the reference
grammar for a given input string. We show how this specialization
grammar can be transformed into an interaction-free form which has the
same practicality as a listing of the individual solutions, but is
produced in less time and space.

The paper proceeds in the following way:
\begin{itemize}
\item Charts can be seen as grammar specializations. The context-free
  case is presented first, then the case of CF-based unification grammars;
\item The rhs of rules can be {\em standardized}: the
  unifications explicitly appearing in the rhs can be standardized to a
  normal form;
\item The notion of a {\em interaction-free}, or {\em IF}, grammar is
  introduced; A unification grammar is called IF when
  its standardized rules have a certain property which guarantees
  absence of conflict when expanding the rhs nonterminals.
\item The chart corresponding to a given input string is generally a
  non-IF grammar. An algorithm which transforms this
  grammar into an equivalent IF grammar is introduced.
\end{itemize}

\section{Charts}

\paragraph{Charts as grammar specializations}

For a CFG in Chomsky Normal Form (binary rules), and for an input
string of length $n$, a chart can be built in time $O(n^3)$ and space
$O(n^2)$ which recognizes whether the string belongs to the
language associated with the grammar. If not only {\em recognition},
but also {\em parsing} of the string is required, then it is
convenient, during the bottom-up construction of the chart, to
associate with each edge the collection of combinations
of daughter edges from which this edge can be obtained. The augmented
chart then requires $O(n^3)$ space, but then each parse tree can be
recovered in a trivial way by starting from the top edge and following
top-down the links between mother and daughter edges. In this way, an
exponential number of parse trees can be represented in polynomial
space.

It has been remarked in \cite{bl89} that an augmented chart for a CFG
$G$, given the input string $\alpha$, can be viewed as a context-free
grammar $G_{\alpha}$, generating only $\alpha$, possibly more than
once. Each mother edge $A$ is seen as a nonterminal, and each
combination of daughter edges $<B,C>$ associated with $A$ is seen as
the rhs $B \: C$ of a rule whose lhs is $A$. This rule corresponds to
a specific instance of some rule of $G$. Each top-down traversal of
$G_{\alpha}$ generates a parse tree for $\alpha$ which is also a parse
tree relative to the full grammar $G$. We will call the grammar
$G_{\alpha}$ a {\em specialization} of $G$ for the given input
string.\footnote{{\bf Charts applied to FSAs} More generally, it is
  possible to directly extend chart-parsing, with the same polynomial
  bounds in time and space, to the situation where the input string of
  words $\alpha$ is generalized to any finite-state
  automaton $FSA$. Chart edges are constructed in the usual way, but
  they now connect automaton nodes rather than positions in the input
  string. The chart constructed over $FSA$ can then be seen as a CFG
  $G_{FSA}$, a specialization of $G$, which generates the {\em intersection}
  of the regular language associated with $FSA$ and the CF language
  associated with $G$ \cite{Lang94}. Thus chart-parsing is
  connected to the closure of context-free languages
  under intersection with a regular language, and the original proof
  of this closure property \cite{barhillel61} can be seen as a
  forerunner (as well as an extension!) of chart-parsing.}

\paragraph{Charts and unification}

Chart-parsing methods extend naturally to unification grammars based
on a context-free backbone, such as LFG \cite{kaplan-bresnan-82} or
DCG \cite{PeWa80}. For ease of exposition, we will use a DCG-like
notation, but we believe the results to be applicable to any CF-based
unification grammar. 

Assuming a grammar with binary branching rules, any grammar rule can
be written in one of the following forms: 
$$R:\myqquad \tt a(A) \ar b(B) \: c(C) \;\; {\cal U}_{\mit R}(A,B,C)$$
for a non-lexical rule $R$, and:
$$S:\myqquad \tt a(A) \ar [t] \;\; {\cal U}_{\mit S}(A)$$
for a lexical rule
$S$. Nonterminals are written as lowercase letters, terminals
under brackets, and uppercase letters are variables representing
feature structures. The notation $\tt {\cal U}_R(A,B,C)$ is an
abbreviation for the set of unification constraints relating the
structures $\tt A$, $\tt B$ and $\tt C$ that appears in rule $R$.

For such a grammar, one can construct a chart/grammar specialization
for a given input string\footnote{Or, more generally, any input FSA.}
in the following way:
\begin{itemize}
\item One constructs a chart for the CF backbone of the grammar as in
  the previous section; This chart can be seen as a specialization of
  the CF-backbone.
\item Each non-lexical rule 
  $$R:\myqquad \tt a \ar b\: c$$
  of the CF-backbone specialization can
  be seen as a specialization of a rule
  $$R':\myqquad \tt a' \ar b'\: c'$$
  of the CF-backbone, where the
  nonterminals $\tt a' ,b',c'$ are specializations of the nonterminals
  $\tt a ,b,c$ (that is, $\tt a'$ is the instance of the nonterminal
  $\tt a$ covering a certain specific substring of the input).
\item Each such rule $R$ is augmented into:
  $$R:\myqquad \tt a(A) \ar b(B) \: c(C) \;\; {\cal U}_{\mit
    R'}(A,B,C)$$
  where $\tt A,B,C$ are fresh variables, and where $\tt
  {\cal U}_{\mit R'}$ is the set of unifications associated with $R'$
  in the original grammar.
\item A similar operation is performed for the lexical rules.
\end{itemize}

The unification grammar obtained by this process is an extension of
the chart/grammar obtained for the CF-backbone. It is a specialization
of the original unification grammar, which is equivalent to this
grammar as far as the given input string is concerned. If one uses the
specialization grammar in a generation mode, expanding nonterminals
top-down and ``collecting'' the unification constraints, one obtains
sets of constraints which collectively describe, {\em when they are
jointly satisfiable}, feature structures associated with the initial
symbol $\tt s$ of the grammar.

The specialization grammar accepts at most the given input string.
Determining whether it {\em does} accept this string can however still be a
difficult problem. Two cases can occur:
\begin{enumerate}
\item The chart/grammar for the CF-backbone contains cycles, that is,
  there exists some nonterminal $A$ (or equivalently, edge) in this
  grammar which calls itself recursively. This can only happen when
  the CF-backbone is an infinitely ambiguous CFG, or, in other words,
  if the given unification grammar is not {\em offline-parsable}
  \cite{PeWa83}; Offline-parsability is guaranteed under certain
  conditions, such as the fact that the CF-backbone does not contain
  any chain production $\tt A \ar B$ or empty production $\tt A \ar
  \epsilon$ \cite{kaplan-bresnan-82}.

  When there are cycles in the
  chart, determining whether there exists a traversal of the
  (unification) specialization grammar which has satisfiable
  constraints is in general undecidable.
\item The full grammar is offline-parsable. Then the chart/grammar for
  the CF-backbone is acyclic. In this case, there are
  only a finite number of top-down traversals of the (unification)
  specialization grammar. For each of the traversals, one can perform
  unification of the collected constraints. Each traversal for which
  the constraints are satisfiable gives rise to a feature structure
  solution for the input string (or, more precisely, to a ``most
  general unifier'' in Prolog terminology, or a ``minimal feature
  structure'' in LFG terminology).
\end{enumerate}

The second case is by far the most important in grammars naturally
occurring in NLP. The recognition problem is then decidable, and all
the feature structure solutions can be enumerated in finite time.
However, there may be an exponential number of these solutions, and it
can be shown that, in general, the problem of determining whether a
solution exists is np-complete in the length of the input string (it
is easy to simulate a boolean satisfaction problem with a non-cyclic
unification grammar). With NL grammars, however, such {\em intrinsic}
complexity apparently does not happen: as was discussed above, NL
grammars are ``almost'' context-free, so that unification features
could in principle be ``almost'' dispensed with, and then, there would
be no unification interactions to cope with.

In the remainder of this paper, we will explore ways to transform
the specialization chart/grammar obtained for an offline-parsable
grammar in such a way that each top-down traversal leads to a
satisfiable set of constraints. Because of the np-completeness
property noted above, such transformed chart/grammars could in theory take
exponential space, but the situation seems to occur rarely, if ever,
with NL grammars.

\section{Standardized rules}\label{Standardized rules} 

\paragraph{Standardized unification sets}

We will use the following notation for a unification constraint:
$$[[A_1,\ldots,A_n],(l_1,B_1),\ldots,(l_m,B_m)],$$
with $1\leq n, 0
\leq m$, and $A_i \not= A_{i'}$ for $i\not=i'$, with the following
interpretation: each $A_i, B_j$ is a variable representing a
feature structure or is a constant representing an atomic feature
structure (such as `sg', `love', etc.); $l_1,\ldots,l_m$ are attribute
labels (such as `subj', `number', etc.); the first element
$[A_1,\ldots,A_n]$ of the constraint, called its {\em identification
  set}, says that the structures $A_1,\ldots,A_n$ should be unified, the
remaining elements $(l_1,B_1),\ldots,(l_m,B_m)$, called {\em access
  relations}, say that the value of the attribute $l_j$ of $A_1$ (and,
therefore, of any $A_i$) is $B_j$.  We will also use the notation
$\top$ for the ``impossible'' unification constraint (which is never
satisfied by any structure). Two simple cases of a unification
constraint are of special interest: the constraint $[[A,A']]$, which
indicates unification of $A$ and $A'$, and the constraint
$[[A],(l,B)]$, which indicates that $B$ is accessed from $A$ through
the attribute $l$.

A finite set of unification constraints is said to be in {\em standardized
  form} if it is the singleton $\set{\top}$ or if it has the following
properties:
\begin{itemize}
\item it does not contain $\top$;
\item the identification sets of the constraints are disjoint;
\item in any given constraint, an attribute label appears at most
  once;
\item in any given constraint, if any of the $A_i$ is a constant, then
  it is the only one, and also $m=0$.
\end{itemize}

A functional structure, in the sense of LFG or related formalisms, can
be seen as an oriented graph whose edges are labeled by attributes,
in such a way that no two edges with the same label originate in the
same node. If one distinguishes a certain ``root'' variable $A$ in a
unification constraint set, then this set can be seen as subsuming all
the functional structures rooted in a node $N_A$ which respect, in the
obvious sense, the description expressed by the constraint set. If the
set is in standardized form and is different from $\top$, then there exist
such structures, and the ``minimal'' functional structure (or, in
Prolog parlance, the most general unifier) which subsumes all these
structures is trivially obtainable from this standardized set. One has the
following property (see \cite{Colmerauer84b} for a slightly different
wording of the result):
\begin{quote}
  If $\cal U$ is a constraint set, then one can obtain in
  linear time (in function of
  the size of $\cal U$) an equivalent standardized set $\cal U'$,
  where ``equivalent'' means: accepting the same functional
  structures. 
\end{quote}

The reduction proceeds by interleaving two steps:
\begin{itemize}
\item If two constraints have non-disjoint identification sets, then
  replace them by a single constraint having as identification set
  the union of the two identification sets, and having as access
  relations the union of the access relations of the two constraints;
\item If a given constraint contains two access relations with the same
  label $(l,B)$ and $(l,C)$, then eliminate the second of these
  relations, and add to the constraint set a new constraint $[[B,C]]$
  expressing the identification of $B$ and $C$.
\end{itemize}
After a finite number of steps, no more such transformations can be done.
At this point, two cases can occur: either some identification set
contains two different atomic constants (which indicates unification
failure) in which case one replaces the whole constraint set by the
singleton $\top$, or this is not so, in which case the unification set
is in standardized form.

\paragraph{Standardized rules}

A grammar rule
$$R:\myqquad \tt a(A) \ar b(B) \: c(C) \;\; {\cal U}_{\mit R}(A,B,C)$$
is said to be standardized if the unification constraint set
${\cal U}_{\mit R}$ is in standardized form. From what has just been said,
it is clear that any grammar rule can be put in standardized form
without altering the structures accepted by the grammar.

\section{Interaction-free grammars}

From any (binary branching) CF grammar $G$, one can obtain a related
unification grammar $G'$, which ``registers'' the derivations of $G$.
If $$\tt a \ar b \: c$$
is a non-lexical rule of $G$, then the
corresponding rule of $G'$ is:
$$\tt a(A) \ar b(B) \: c(C), \; [[A],(l,B),(r,C)],$$
where $\tt
A,B,C$ are fresh variables, and where the constraint expresses that
the left constituent of $\tt A$ is $\tt B$, its right constituent $\tt
C$. Similarly, for a lexical rule: $$\tt a \ar [t]$$
of $G$, the corresponding rule
of $G'$ is:
$$\tt a(A) \ar [t], \; [[A],(lex,t)],$$
which indicates that the lexical value of $A$ is $\tt t$.

The grammar $G'$, which we call a a {\em pure derivation grammar},
accepts the same strings as $G$, and assigns to them a functional
description which is just an encoding of the derivation tree for $G$.

It is clear that, in any top-down traversal of $G'$, the constraints
cannot clash. In fact, if one considers the set of constraints
collected during such a traversal, it is obvious that this set is a
{\em standardized} unification set, so that no interaction exists between
the different variables.

We now introduce a definition which generalizes the situation obtained
with $G'$:
\begin{quote}
  A grammar is called an {\em interaction-free}, or {\em IF}, grammar,
  if all its standardized rules are interaction-free, that is, have
  the following two properties:
  \begin{itemize}
    \item the unification set of the rule is not $\set{\top}$;
    \item if $\tt B$ is the variable associated with any rhs nonterminal 
    $\tt  b$, then this variable does not appear in the identification
    set of any unification constraint in the rule.
  \end{itemize}
\end{quote}

It can be checked that this condition is respected by grammar $G'$. In
an interaction-free grammar, any top-down traversal gives rise to a
standardized set of unifications, so that no clash can occur between
constraints.

\subparagraph{Standardized unification sets and interaction-free rules}
The choice of representation for unification constraints made in
section \ref{Standardized rules} was obviously not the only one
possible. A more standard, but equally expressive, notation for
unifications would have been:
$$(A,l,B)$$
for access constraints, and:
$$C=D$$
for equality constraints.

The interest of the notation of section \ref{Standardized rules} is
that the notion of standardization can be defined in it
straighforwardly, and that this last notion is directly related to the
property of interaction-freeness. Because of the form of a
standardized rule, it is obvious that a variable $B$ that does not
appear in the identication set of any unification constraint can be
set arbitrarily by the ``nonterminal call'' $b(B)$, without risk of
clashing with variables set by other nonterminal calls in the rule.

This is to be contrasted with the situation had the more standard
notation been used. Consider, in such a notation, the following rule:
$$\tt a(A) \ar b(B) \: c(C), \; (A,l1,B), (A,l1,D), (D,l2,C).$$

In this rule there can be a conflict between the assignments given to
$\tt B$ and to $\tt C$. This is because, implicitly, the value of the
$\tt l2$ attribute in $\tt B$ is equal to $\tt C$. On the other hand,
using the notation of section \ref{Standardized rules}, the
standardized rule is written as
$$\tt a(A) \ar b(B) \: c(C), \; [[A],(l1,B)], [[B,D],(l2,C)]$$
which is immediately seen {\em not} to be interaction-free: $\tt B$ appears
in the identification set of the second constraint.

\section{Making an acyclic grammar interaction-free}

Let us consider the chart/grammar specialization $G_\alpha$ of an
offline-parsable grammar $G$ for a string $\alpha$. This grammar is
acyclic, that is, no nonterminal calls itself recursively. We will now
introduce an algorithm for transforming $G_\alpha$ into an equivalent
acyclic IF grammar. We will suppose that the rules of 
$G_\alpha$ have been standardized.

The algorithm works by iteratively replacing non-IF rules of the
grammar $G_\alpha$ by equivalent, ``more IF'', rules, until no
non-IF rule remains in the grammar:
\begin{enumerate}
\item We may suppose that there exists a non-IF rule in the
  grammar, otherwise we are finished. Let us say that a given rule
  $R'$ is {\em below} another (different) rule $R''$ if there is a
  sequence of rules $R_1=R'',R_2,\ldots,R_n=R'$ in the grammar such
  that $R_{i}$ contains in its rhs a nonterminal which appears on the
  lhs of $R_{i+1}$. Because the grammar is acyclic, there must be some
  non-IF rule $R$ such that any rule below it is IF; otherwise
  there would be a non-IF rule which is below itself, which would
  imply grammar cyclicity.
\item Take $R$. There exists a nonterminal $\tt b(B)$ in its rhs such that
  $\tt B$ appears in the identification set of some constraint of
  $R$. Partially evaluate $\tt b(B)$ by replacing it with the right-hand
  sides of the (IF) rules which define $\tt b$.
\item This may produce several different ``copies'' of $R$, or none,
  depending on the number of rules which define $\tt b$ in the grammar. In
  the latter case, no copy is produced, so that $R$ has disappeared
  (it was unproductive, that is, could not generate anything.)
\item Is some copies of $R$ are produced, standardize them. If, after
  standardization, the constraint set of the rule is $\set{\top}$,
  eliminate the corresponding rule (it is unproductive).
\item Go to step 1.
\end{enumerate}
The algorithm terminates after a finite number of steps, because in an
acyclic grammar, partial evaluation can only be performed a finite number
of times.

Also, one can see, by a simple induction, that the IF grammar
obtained only contains productive nonterminals, that is,
nonterminals which do generate some string with satisfiable
constraints.

Let us consider a simple example. Suppose that the input string is
``john read here'', and that the chart/grammar specialization is:
\vspace*{-1ex}\begin{alltt} \tt 
s(S) \armath np(NP) vp(VP), [[S],(l,NP),(r,VP)], 
  [[NP],(n,X)], [[VP],(n,X)]
vp(VP) \armath v(V) a(A), [[VP],(l,V),(r,A),(n,Y)],
  [[V],(n,Y)]
v(V) \armath [read], [[V],(lex,read),(n,sg)]
v(V) \armath [read], [[V],(lex,read),(n,pl)]
np(NP) \armath [john], [[NP],(lex,john),(n,sg)]
a(A) \armath [here], [[A],(lex,here)]
\end{alltt}
where $n$ refers to the attribute `number'.
All rules are IF, apart from the $\tt s$ and $\tt vp$ rules. All the rules
below the $\tt vp$ rules are IF. We evaluate partially $\tt v(V)$ in this
rule, which gives the two rules:
\vspace*{-1ex}\begin{alltt} \tt 
vp(VP) \armath [read] a(A), [[V],(lex,read),(n,sg)],
   [[VP],(l,V),(r,A),(n,Y)], [[V],(n,Y)]
vp(VP) \armath [read] a(A), [[V],(lex,read),(n,pl)],
   [[VP],(l,V),(r,A),(n,Y)], [[V],(n,Y)]
\vspace*{1ex}\end{alltt}
or, after standardization:
\vspace*{-1ex}\begin{alltt} \tt 
vp(VP) \armath [read] a(A), [[V],(lex,read),(n,sg)],
   [[VP],(l,V),(r,A),(n,Y)], [[sg,Y]]
vp(VP) \armath [read] a(A), [[V],(lex,read),(n,pl)],
   [[VP],(l,V),(r,A),(n,Y)], [[pl,Y]]
\vspace*{1ex}\end{alltt}
These rules are IF, because the nonterminal $\tt a(A)$ does not appear
in any identification set. The only non-IF rule is now the rule
for $\tt s$, and we partially evaluate $\tt np(NP)$ in this rule, giving, after
standardization:
\vspace*{-1ex}\begin{alltt} \tt 
s(S) \armath [john] vp(VP), [[NP],(lex,john),(n,sg)], 
  [[S],(l,NP),(r,VP)], [[VP],(n,X)], [[X,sg]]
\vspace*{1ex}\end{alltt}
This rule is again non-IF. We partially evaluate $\tt vp(VP)$, giving the two rules:
%
%
\vspace*{-1ex}\begin{alltt} \tt 
s(S) \armath [john] [read] a(A), [[V],(lex,read),(n,sg)], 
  [[VP],(l,V),(r,A),(n,Y)], [[sg,Y]], [[NP],(lex,john),(n,sg)],
  [[S],(l,NP),(r,VP)], [[VP],(n,X)], [[X,sg]]
s(S) \armath [john] [read] a(A), [[V],(lex,read),(n,pl)],
  [[VP],(l,V),(r,A),(n,Y)], [[pl,Y]], [[NP],(lex,john),(n,sg)],
  [[S],(l,NP),(r,VP)], [[VP],(n,X)], [[X,sg]]
\vspace*{1ex}\end{alltt}
which, during standardization, become:
\vspace*{-1ex}\begin{alltt} \tt 
s(S) \armath [john] [read] a(A), [[V],(lex,read),(n,sg)], 
  [[VP],(l,V),(r,A),(n,Y)], [[NP],(lex,john),(n,sg)], 
  [[S],(l,NP),(r,VP)], [[Y,X,sg]]
s(S) \armath [john] [read] a(A), [[V],(lex,read),(n,pl)],
  [[VP],(l,V),(r,A),(n,Y)], [[NP],(lex,john),(n,sg)], 
  [[S],(l,NP),(r,VP)], [[pl,Y,X,sg]]
\vspace*{1ex}\end{alltt}
In the second rule, the constraint $\tt [[pl,Y,X,sg]]$ reduces to $\top$,
and so the rule is eliminated. As for the first rule, it is is already
standardized, as well as IF. The grammar obtained is now IF.

It should be noted that, in the IF rule obtained for $\tt s$, the
adjunct $\tt a(A)$ is now ``free'': because $\tt A$ does not appear in any
identification set of the rhs, whatever the values $\tt A$ may take, it
won't interact anymore with the rule. Thus, even if there were
many analyses for the adjunct, this unique rule would take care of
all of them. This is to be contrasted with the case where we would
have evaluated all the nonterminals appearing in the rhs of the $\tt s$
rule, where each solution for the adjunct would have been explicitly
represented along with the others.

The example, although simple, displays the crucial feature of the
transformation into IF form: partial evaluation is performed only in
case of need; as soon as a nonterminal can no longer interact with its
siblings in a rule, it is not expanded further. If this nonterminal
itself has a number of possible solutions, these solutions are kept
``factorized'' in the grammar.

\paragraph{Ambiguous structures seen as IF grammars}

Grammars are a kind of and/or representation: alternative rules for
expanding the same nonterminal correspond to a disjunction, while a
string of nonterminals on the rhs of a given rule corresponds to a
conjunction. 

An acyclic IF grammar with no unproductive nonterminals is an
efficient representation of the several feature structures it
generates: if one traverses this grammar top-down from its initial
nonterminal $s$, then one never backtracks, because all nonterminals
are productive, and because the collected constraints cannot clash.
Thus one can enumerate all the structures in a direct way. Such
grammar representations can be likened to finite-state
representations for dictionaries, which, although they are more
compact than simple lists of words, have for most purposes the same
practicality as such lists.

\section{Conclusion}

This paper has brought together two current lines of research: (i)
viewing charts as grammar specializations, and (ii) extending
chart-like ambiguity packing to unification grammars. By doing so, it
has clarified the principles underlying the nature of shared
representations, stressed the link between grammars and
representations, and opened the way for further applications in
computational linguistics.

\paragraph{Acknowledgments} Thanks for discussions and comments to 
Max Copperman, Lauri Karttunen, Bernard Lang, John Maxwell and Eric
Villemonte de la Clergerie.

\bibliographystyle{plain}

\end{document}